\documentclass[onecolumn,ournal]{IEEEtran}
\IEEEoverridecommandlockouts
\usepackage{amsmath,amssymb,amsfonts}
\usepackage{booktabs}

\usepackage{algorithmic}
\usepackage{graphicx}
\usepackage{subfig}
\usepackage{epstopdf}
\usepackage{textcomp}
\usepackage{xcolor}
\usepackage{cuted}

\def\BibTeX{{\rm B\kern-.05em{\sc i\kern-.025em b}\kern-.08em
    T\kern-.1667em\lower.7ex\hbox{E}\kern-.125emX}}
\begin{document}

\title{Study of Novel Sparse Array Design Based on the Maximum Inter-Element Spacing Criterion
\\
\author{Wanlu Shi, and Yingsong Li, \IEEEmembership{Senior Member, IEEE},~Rodrigo C. de Lamare,~\IEEEmembership{Senior Member,~IEEE} \vspace{-1.5em}}
\thanks{}
\thanks{Wanlu Shi is with the College of Information and Communication Engineering, Harbin Engineering University, Harbin 150001, China.}
\thanks{Yingsong Li is with the Key Laboratory of Intelligent Computing and Signal Processing
Ministry of Education, Anhui University, Hefei, Anhui, China (e-mail: liyingsong@ieee.org){\it (Corresponding author: Yingsong Li)}}
\thanks{R. C. de Lamare is with the Centre for Telecommunications Research (CETUC),
Pontifical Catholic University of Rio de Janeiro (PUC-Rio),
G¨¢vea, Rio de Janeiro - Brazil and the University of York, UK.}
}

\maketitle
\begin{abstract}
A novel sparse array (SA) structure is proposed based on the maximum inter-element spacing (IES) constraint (MISC) criterion. Compared with the traditional MISC array, the proposed SA configurations, termed as improved MISC (IMISC) has significantly increased uniform degrees of freedom (uDOF) and reduced mutual coupling. In particular, the IMISC arrays are composed of six uniform linear arrays (ULAs), which can be determined by an IES set. The IES set is constrained by two parameters, namely the maximum IES and the number of sensors. The uDOF of the IMISC arrays is derived and the weight function of the IMISC arrays is analyzed as well. The proposed IMISC arrays have a great advantage in terms of uDOF against the existing SAs, while their mutual coupling remains at a low level. Simulations are carried out to demonstrate the advantages of the IMISC arrays.
\end{abstract}
\begin{IEEEkeywords}
Uniform degrees of freedom, sparse array (SA), mutual coupling, maximum inter-element spacing constraint, direction-of-arrival (DOA) estimation.
\end{IEEEkeywords}

\section{Introduction}
There is a growing body of literature that emphasizes the vital role of sensor array techniques in diverse applications, such as seismology~\cite{optimum_array_processing}, radar~\cite{radar}, sonar~\cite{sonar}, and others. {An important problem in sensor array technique is direction finding~\cite{music}, especially the case when there are more targets than the number of sensors~\cite{optimum_array_processing,fourth_order}.} In this context, the difference coarray (DCA) principle is one of the feasible choices to produce plenty uniform degrees of freedom (uDOF), {which makes possible to use fewer sensors estimate more targets~\cite{coarray,MRA,MHA}.} Based on the DCA concept, many classical sparse arrays (SAs) have been introduced, among which the nested arrays (NAs)~\cite{NESTED} and coprime arrays (CAs) are very popular~\cite{coprimeC,COPRIME,Extended_coprime,stap_coprime,listomp}. The NAs can achieve $O(N^2/2)$ uDOF using $N$ sensors but are sensitive to the mutual coupling (MC) among sensors. Unlike NAs, CAs are robust to MC, but their uDOF are less than those of NAs~\cite{coprimeC,COPRIME}.
\par
Many existing SAs have been developed via the prototype NAs and CAs ~\cite{improved_nested,CADIS,ANA,TSONA,Extended_coprime}. Early SAs have focused on increasing the achievable uDOF. To this end, the improved NA (INA)~\cite{improved_nested}, the CA with displaced subarrays (CADiS)~\cite{CADIS} and the two side-extended nested array (TSENA)~\cite{TSONA} are proposed. These SAs show considerable improvement on uDOF. However, with the growing of electromagnetic equipments, MC is inevitable and needs to be considered when designing SAs. In this regard, several SAs have been proposed based on NAs and CAs. For example, the super nested arrays (SNAs)~\cite{SNAC1,SNAC2,SNA1,SNA2}, the extended padded coprime arrays (ePCAs)~\cite{padded_coprime} and the improved coprime nested arrays (ICNA)~\cite{ICNA} are examples of SAs designed based on NAs and CAs. It is worth noting that the ICNA realizes $O( 4Q^2/7)$ uDOFs with $Q$ sensor elements and MC among sensors is very low.
Another general SA design method is the ULA fitting scheme, based on which the ULA fitting with 4 base layer (UF-4BL) has been proposed with low MC~\cite{UF,UF_ICASSP}.
Beyond the aforementioned SA design approaches, the maximum inter-element spacing (IES) constraint (MISC) criterion is also promising~\cite{MISC}. The MISC criterion aims to design SAs with an inter-element spacing set, which is constrained by the maximum element spacing and the number of sensors. The MISC arrays  provide a good balance between MC and uDOF.

In this work, we develop an improved MISC (IMISC) array based on the MISC approach with an increased uDOF and reduced MC as compared to the traditional MISC array. In particular, we carry out an analysis that shows that IMISC arrays reach $O( 2Q^2/3)$ uDOF, which is significantly higher than those uDOF achieved by existing sparse arrays. Moreover, the MC of the proposed IMISC arrays remains at a reduced level. Simulations illustrate the excellent performance of the proposed IMISC arrays.

\section{Coarray Concept and Mutual Coupling}\label{fundamental}
\subsection{Difference Coarray Model}

Let us assume an $Q$-sensor linear array with location set
\begin{equation}
\mathbb{L} = \left\{ {{p_0},p_1, p_2,\dots,p_{Q-1}} \right\},
\label{position_set}
\end{equation}
then the associated steering vector can be obtained as $\mathbf{v}_{\phi}=\left[e^{j \frac {2\pi}{\lambda} p_0/ \sin(\phi)},\cdots,e^{j \frac {2\pi}{\lambda} p_{Q-1}/ \sin(\phi)}\right]^T$, where $\phi$ is the signal impinging direction. Then imagine $R$ uncorrelated far-field narrowband signals impinging from directions $\left\{\phi_i,i=1,\cdots,R\right\}$ with powers $\left\{\sigma^2_i,i=1,\cdots,R\right\}$. In this regard, the observed signal can be expressed as
\begin{equation}
{\mathbf{x}} = {\mathbf{V}}{\mathbf{s}} + {\mathbf{n}},
\label{received_data_one_snapshot}
\end{equation}
where $\mathbf{V}\triangleq\left[\mathbf{v}_{\phi_1},\cdots,\mathbf{v}_{\phi_R}\right]$ is the array manifold matrix, $\mathbf{s} \triangleq \left[\mathbf{s}_1,\cdots,\mathbf{s}_R\right]^T$ denotes the signal vector, ${\mathbf{n}}$ is the white Gaussian noise with zero mean. Then, the covariance matrix of (\ref{received_data_one_snapshot}) can be calculated as
\begin{equation}
{\mathbf{R}_{\mathbf{x}}} \triangleq E[\mathbf{x}{\mathbf{x}^H}] = \mathbf{V}\mathbf{R}_{\mathbf{s}}{\mathbf{V}^H} + \sigma_n^2 \mathbf{I},
\label{covariance_matrix}
\end{equation}
where $\sigma ^2_n$ is the power of noise and $\mathbf{R}_{\mathbf{s}}$ represents the signal covariance matrix. In this case, the vectorized version of~(\ref{covariance_matrix}) can be obtained as
\begin{equation}
{\mathbf{w}}\triangleq{\rm{vec}}({\mathbf{R}_{\mathbf{x}}}) = ({\mathbf{V}}^*\odot{\mathbf{V}}){\mathbf{q}}+ \sigma_n^2 {\mathbf{1}}_n,
\label{vectorized_covariance_matrix}
\end{equation}
where ${\mathbf{1}}_n={\rm{vec}}({\mathbf{I}}_N)$, ${\mathbf{q}} = \left[\sigma ^2_1,\cdots,\sigma ^2_R\right]^T$, and $\odot$ means the Khatri-Rao product. The DCA concept originates from~(\ref{vectorized_covariance_matrix}), where ${\mathbf{q}}$ serves as the observed signal of the DCA. Based on~(\ref{vectorized_covariance_matrix}), the DCA of array $\mathbb{L}$ has the following position set~\cite{NESTED}
\begin{equation}
\mathbb{D}= \left\{ {{p_a-p_b},a,b = 0,1, \cdots Q-1} \right\}.
\label{DCA}
\end{equation}
\subsection{Coupling Leakage}
In practice, the MC among sensors should be considered. To this end, the MC matrix $\mathbf{A}$ can be interpolated into~(\ref{received_data_one_snapshot}), resulting in
\begin{equation}
{\mathbf{x}} = {\mathbf{A}} {\mathbf{V}}{\mathbf{s}} + {\mathbf{n}},
\label{received_data_one_snapshot_with_coupling}
\end{equation}
Based on~\cite{MISC,coupling1,coupling3}, the MC matrix for linear arrays can be formulated as {a $D$-banded matrix given by}
\begin{equation}
{\mathbf{A}}_{b,c}=\left\{{\begin{array}{*{20}{l}}
a_{|p_b-p_c|},&|p_b-p_c|\le D,\\
0,&{\rm{elsewhere}},
\end{array}} \right.
\label{c_approximate}
\end{equation}
where $p_b,p_c\in \mathbb{L}$ and $a_d,d\in [0,D]$ represent elements in $\mathbf{A}$ that satisfy
\begin{equation}
\left\{{\begin{array}{*{20}{l}}
a_0=1\textgreater |a_1|\textgreater |a_2|\textgreater\cdots\textgreater|a_D|,\\
|a_i/a_j|=j/i,\quad i,j\in [{\color{red}1},D].
\end{array}} \right.
\label{c_coefficients}
\end{equation}
Coupling leakage is a parameter to evaluate the MC effect, which is expressed as~\cite{SNA1,ANA,MISC}
\begin{equation}
E=\frac {||{\mathbf{A}}-{\rm{diag}}({\mathbf{A}})||_F} {||{\mathbf{A}}||_F}.
\label{coupling_leakage}
\end{equation}
Typically, higher coupling leakage means larger MC. Moreover, equations~(\ref{c_approximate}),~(\ref{c_coefficients}) and~(\ref{coupling_leakage}) reveal the relationship between MC and the weight function $w(n)$. For linear arrays, the {weight function $w(n)$ indicates} the number of element pairs {whose} IES {is} $n$. {For SA $\mathbb{L}$, $w(n)$ can be obtained as~\cite{MISC}
\begin{equation}\nonumber
w(n)=|\{ (p_i,p_j)|p_i-p_j = n;p_i,p_j \in \mathbb{L}\}|.
\end{equation}
Based on the analysis above,} it is noted that $w(1)$, $w(2)$ and $w(3)$ dominate the MC.

\section{Proposed IMISC array}
In this section, the IMISC array is developed, which is devised based on the MISC criterion{~\cite{MISC}}. {Compared with the traditional MISC array, the proposed IMISC SA has higher uDOF and lower coupling leakage.} The IMISC SA has key merits. First, IMISC has a closed-form expression for sensor location and uDOF. Moreover, compared with the existing SAs, IMISC has higher uDOF and provides better {\color{red}a} balance between uDOF and MC.

\begin{figure}[htp]
\centering
\centerline{\includegraphics[width=1\columnwidth,height=1.4cm]{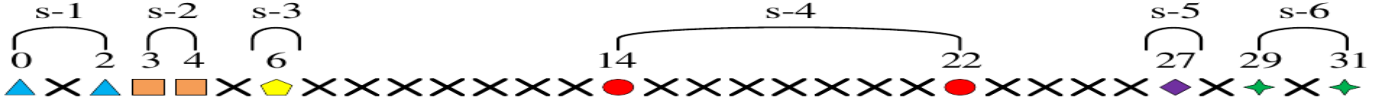}}
\caption{IMISC array structure with~$Q=10$ and $M=8$.}
\label{IMISC_STRUCTURE}
\end{figure}

The IMISC array can be identified using an IES set for a given sensor amount $Q$. Particularly, the IES set is expressed as $\mathbb{S}$ and the maximum IES is denoted as $M$. In this regard, $\mathbb{S}$ for IMISC can be expressed as
\begin{equation}
\mathbb{S}_{\text{IMISC}}=\left\{ \begin{array}{l}
\underbrace {2,...,2}_{\frac{M}{4} - 1},1,1,\frac{M}{2} - 2,\\
\underbrace {\frac{M}{2} - 1,...,\frac{M}{2} - 1}_{\frac{M}{4} - 2},\underbrace {M,...,M}_{Q - M},\\
\frac{M}{2} + 1,\underbrace {\frac{M}{2} + 1,...,\frac{M}{2} + 1}_{\frac{M}{4} - 2},2,\underbrace {2,...,2}_{\frac{M}{4} - 1},
\end{array} \right\}
\label{spacing}
\end{equation}
where
\begin{equation}
M = 4\lfloor \frac{Q+2}{6}\rfloor,Q\ge10,\\
\label{M}
\end{equation}
where $\lfloor \cdot \rfloor$ is the floor operator. The location set associated to~(\ref{spacing}) is given as
\begin{equation}
\begin{aligned}
&\mathbb{L}_{\text{IMISC}} =\\
\small&\left\{ \begin{array}{l}
\underbrace {0,...,\frac{M}{2} - 2}_{{\text{ULA 1, IES=2}}},\underbrace {\frac{M}{2} - 1,\frac{M}{2}}_{{\text{ULA 2, IES=1}}},\underbrace {M - 2,...,\frac{{{M^2}}}{8} - \frac{M}{4}}_{{\text{ULA 3, IES=}}\frac{M}{2} - 1},\\
\underbrace {\frac{{{M^2}}}{8} + \frac{{3M}}{4},...,MQ - \frac{{7{M^2}}}{8} - \frac{M}{4}}_{{\text{ULA 4, IES=}}M},\\
\underbrace {MQ - \frac{{7{M^2}}}{8} + \frac{M}{4} + 1,...,MQ - \frac{{3{M^2}}}{4} - \frac{M}{2} - 1}_{{\text{ULA 5, IES=}}\frac{M}{2}+1},\\
\underbrace {MQ - \frac{{3{M^2}}}{4} - \frac{M}{2} + 1,...,MQ - \frac{{3{M^2}}}{4} - 1}_{{\text{ULA 6, IES=2}}}.
\end{array} \right\}
\end{aligned}
\label{Structure}
\end{equation}
One can tell from~(\ref{Structure}) that IMISC is composed of 6 sub-ULAs and a specific array configuration of IMISC is provided in~Fig.~\ref{IMISC_STRUCTURE}, where ${\text{s-}}i$, $i=1,...,6$ represent sub-ULA~1,...,sub-ULA~6, respectively.

\section{Analysis of IMISC}
In this section, we carry out an analysis of IMISC arrays to obtain the achievable uDOF and their weight functions.
\subsection{Achievable uDOF}
Defining $\mathbb{D}_{\text{IMISC}}$ as the DCA of the IMISC, then the consecutive part of $\mathbb{D}_{\text{IMISC}}$ can be expressed as
\begin{equation}
\mathbb{C}_{\text{IMISC}}=[-MQ + \frac{{3{M^2}}}{4} + \frac{M}{2} - 1,MQ - \frac{{3{M^2}}}{4} - \frac{M}{2} + 1].
\label{consecutive_part1}
\end{equation}
The proof of~(\ref{consecutive_part1}) can be found in the Appendix. Based on~(\ref{consecutive_part1}), the final uDOF of IMISC is
\begin{equation}
{\text{uDOF}}_{\text{IMISC}}=2MQ - \frac{{3{M^2}}}{2} - M + 3.
\label{uDOF1}
\end{equation}
Substituting~(\ref{M}) to~(\ref{uDOF1}) and omitting the floor operator, the uDOF of IMISC can be approximately obtained as
\begin{equation}
{\text{uDOF}}_{\text{IMISC}}=\frac{{{\rm{2}}{Q^2}}}{3} - \frac{{2Q}}{3} - 1.
\label{uDOF2}
\end{equation}
Considering the fact that $M$ and $Q$ are integers, the accurate expression for ${\text{uDOF}}_{\text{IMISC}}$ is
\begin{equation}
\begin{aligned}
{\text{uDOF}}_{\text{IMISC}}=\left\{ {\begin{array}{*{20}{l}}
\frac{{{\rm{2}}{Q^2}}}{3} - \frac{{2Q}}{3} - 1,&Q\% 6 = 4,3,\\
\frac{{{\rm{2}}{Q^2}}}{3} - \frac{{2Q}}{3} + \frac{5}{3},&Q\% 6 = 5,2,\\
\frac{{{\rm{2}}{Q^2}}}{3} - \frac{{2Q}}{3} + 3,&Q\% 6 = 0,1,
\end{array}} \right.
\label{uDOF3}
\end{aligned}
\end{equation}
where $\%$ is the remainder operator. Based on~(\ref{uDOF2}) and~(\ref{uDOF3}), one can find that IMISC can use $O(Q)$ sensors {to} produce $O( 2Q^2/3)$ uDOF, which is higher than most existing SAs. For comparison, the uDOF for the traditional MISC array is provided here
{\begin{equation}
\begin{aligned}
{\text{uDOF}}_{\text{MISC}}=\left\{ {\begin{array}{*{20}{l}}
\frac{{{Q^2}}}{2} + 3Q - 8.5,&Q\% 4 = 1,\\
\frac{{{Q^2}}}{2} + 3Q - 9,&Q\% 2 = 0,\\
\frac{{{Q^2}}}{2} + 3Q - 10.5,&Q\% 4 = 3.
\end{array}} \right.
\end{aligned}
\label{uDOF_MISC}
\end{equation}}
From~(\ref{uDOF3}) and~(\ref{uDOF_MISC}), the IMISC has a higher uDOF than traditional MISC.
\subsection{Weight Function}
Based on~(\ref{spacing}), the first three values in the weight function can be obtained as
\begin{equation}
\begin{aligned}
&w(1)=2, \quad w(2)=\left\{ {\begin{array}{*{20}{l}}
2\lfloor\frac{Q+2}{6}\rfloor,&Q\ge 16,\\
5,&16>Q\ge10,
\end{array}} \right.\\
&\quad \quad \quad \quad w(3)=\left\{ {\begin{array}{*{20}{l}}
1,&Q\ge 16,\\
2,&16>Q\ge10.
\end{array}} \right.
\label{w1}
\end{aligned}
\end{equation}
For comparison, the first three values in the weight function for MISC are given by
\begin{equation}
\begin{aligned}
w(1)=1,w(2)=2\lfloor\frac{Q}{4}\rfloor-3,
w(3)=\left\{ {\begin{array}{*{20}{l}}
1,&Q\neq 9,\\
2,&Q=9.
\end{array}} \right.
\label{w2}
\end{aligned}
\end{equation}
Based on~(\ref{w1}) and~(\ref{w2}), it is obvious that the IMISC has higher $w(1)$ but lower $w(2)$ than MISC. Therefore, when $Q$ is sufficiently large, the coupling leakage of the IMISC will lower than the traditional MISC.
\section{Numerical Simulations}\label{NE}
In this section, the proposed IMISC array is compared with the Co-prime ~\cite{Extended_coprime}, MISC~\cite{MISC}, SNA~\cite{SNA1,SNA2}, TSONA~\cite{TSONA}, ePCA~\cite{padded_coprime}, ICNA~\cite{ICNA} and UF-4BL~\cite{UF} arrays.
DOA estimations are carried out utilizing the spatial smoothing MUSIC algorithm, and the number of snapshots is set to 1000~\cite{NESTED}. In all examples, coupling parameters are chosen according to~$a_i=a_1e^{-j(i-1)\pi/8}/i, \, i=2,\dots,100$. In this regard, the larger $|a_1|$ indicates higher MC effect. Other parameter estimation algorithms that exploit sparsity can also be considered \cite{intadap,jidf,sjidf,rrdoa}.

\subsection{uDOF and Coupling Leakage}

\begin{figure}[hpt]
\centering
    \subfloat[uDOF.]{\includegraphics[width=0.25\columnwidth,height=3.5cm]{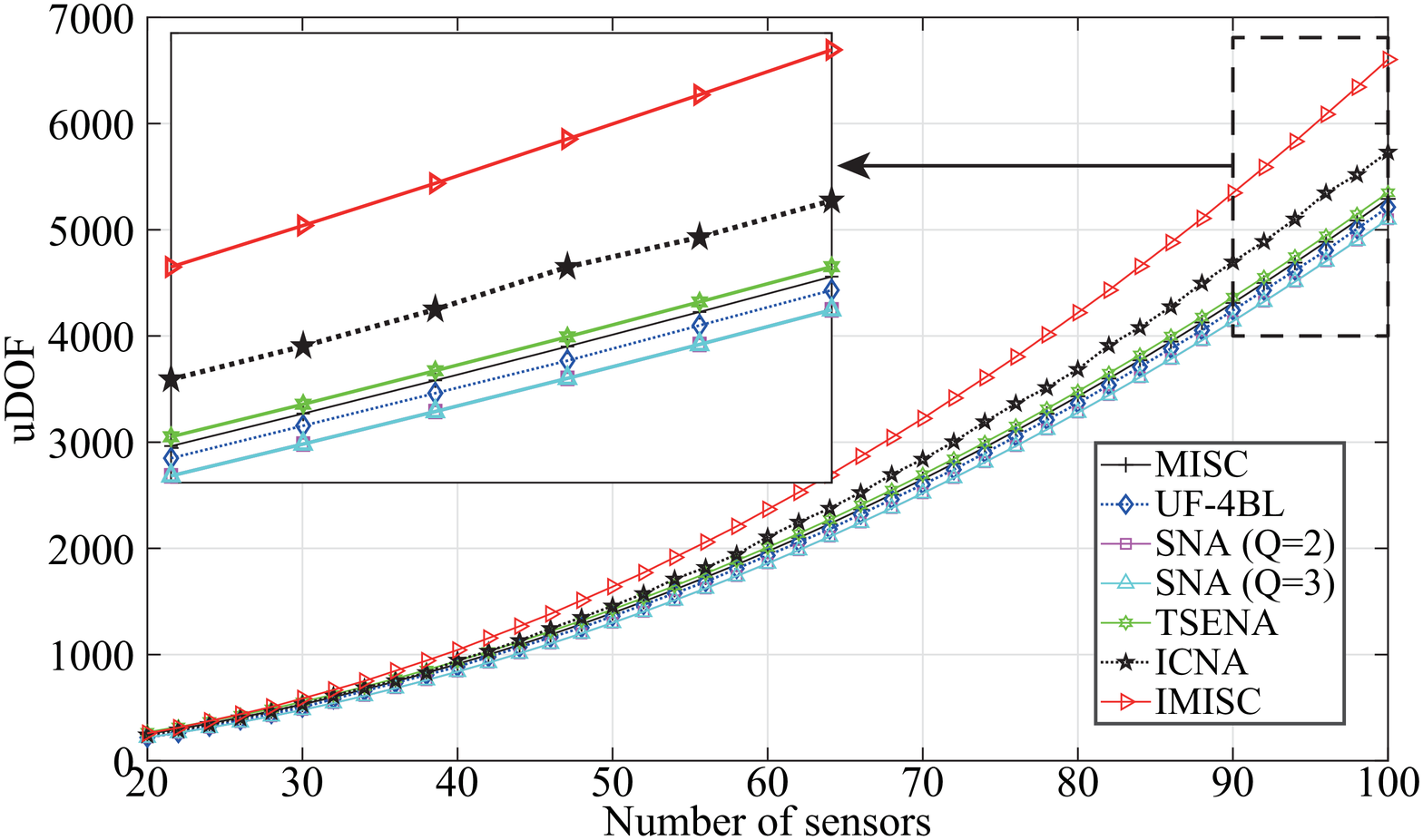} }
    \subfloat[Coupling leakage.]{\includegraphics[width=0.5\columnwidth,height=3.5cm]{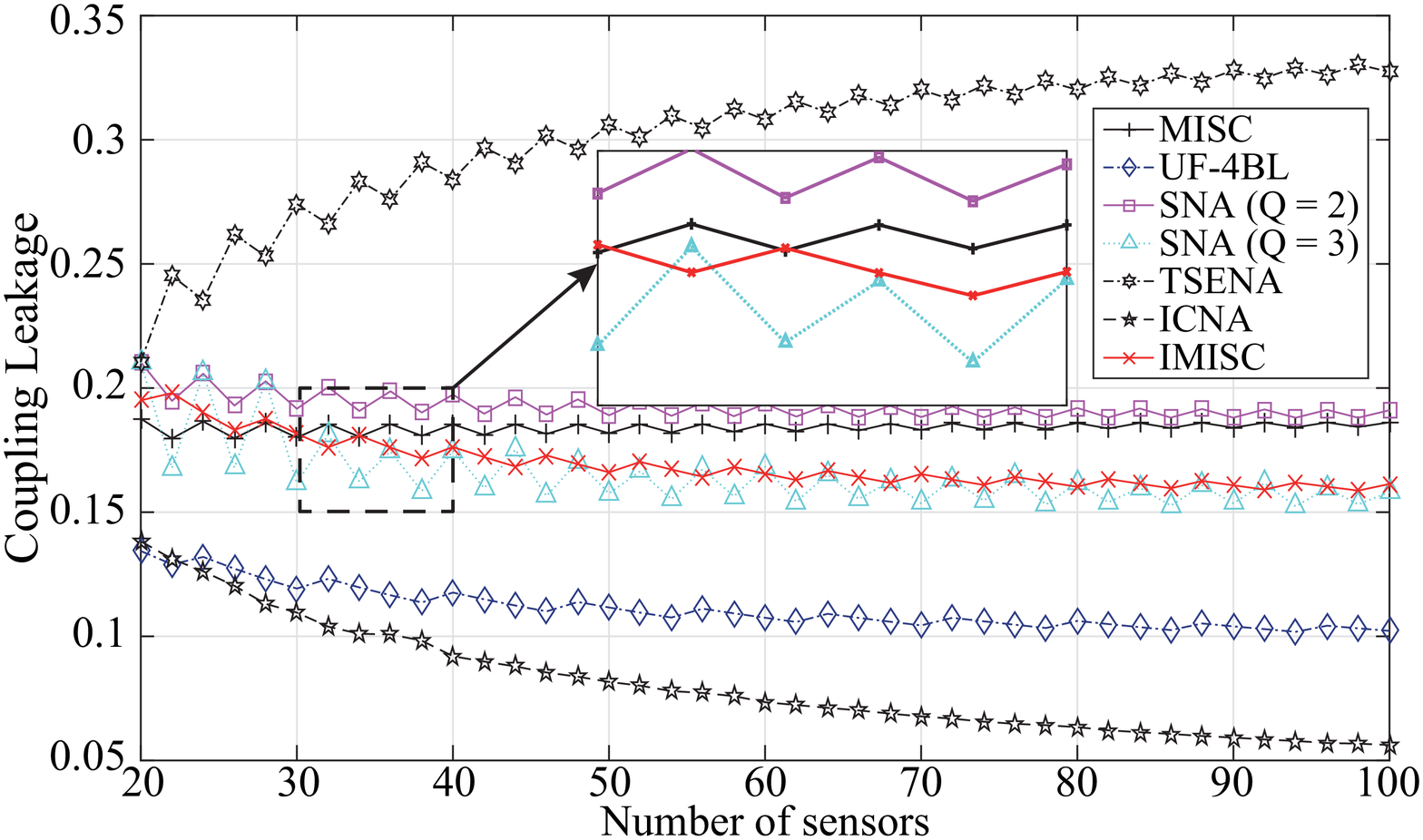} }
    \caption{uDOF and coupling leakage versus number of sensors.}
    \label{uDOF-CL}
\end{figure}

In the first example, the uDOF and coupling leakage versus number of sensors are presented, respectively, where the number of sensors vary from 20~to~100. Fig.~\ref{uDOF-CL}(a) shows the variation of the uDOF when the number of sensors is altered. It is clear that the proposed IMISC array has the highest uDOF among the relevant SA configurations. Besides, with the increasing of the number of sensors, the advantage of IMISC on uDOF is increased as well. This is due to the fact that the IMISC reaches $O(2Q^2/3)$ uDOF, while the maximum uDOF for other SAs is $O(4Q^2/7)$.
\par
Fig.~\ref{uDOF-CL}(b) shows the coupling leakage versus the number of sensors. One can see that the couple leakage for IMISC is lower than that of the traditional MISC array but higher than ICNA and UF-4BL. Therefore, based on the results shown in~Fig.~\ref{uDOF-CL}, we claim that the proposed IMISC array provides better balance between uDOF and coupling leakage than existing SAs.
\vspace{-3mm}
\subsection{Root-Mean-Square Error (RMSE) Performance with various SNR, $|a_1|$, and snapshots}

In the second example, the RMSE performance in different conditions is considered. All the RMSE results are obtained through~500 trials.
\begin{figure}[hpt]
\centering
    \subfloat[RMSE versus SNR.]{\includegraphics[width=0.25\columnwidth,height=3.3cm]{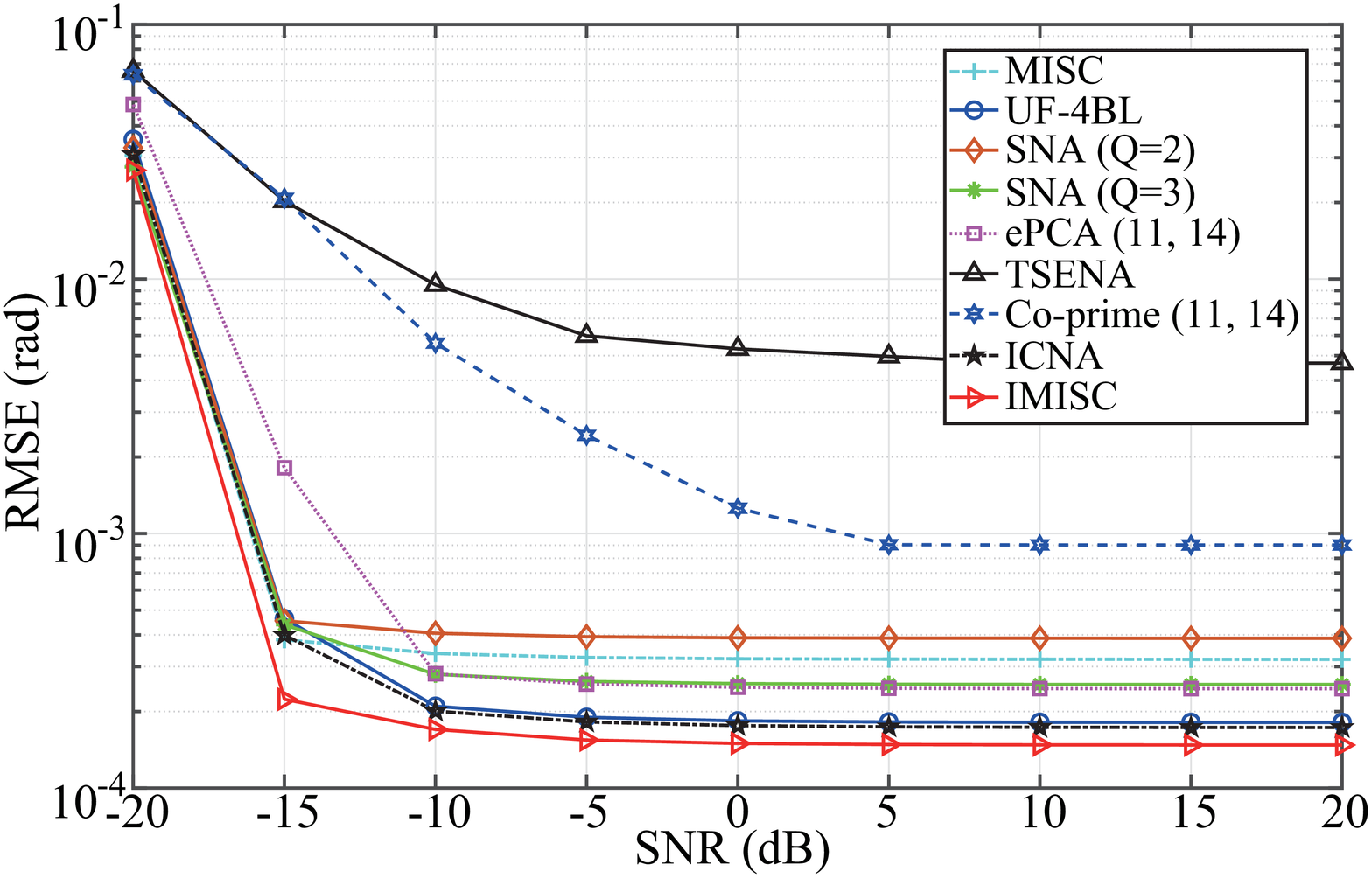} }
    \subfloat[RMSE versus $|a_1|$.]{\includegraphics[width=0.5\columnwidth,height=3.3cm]{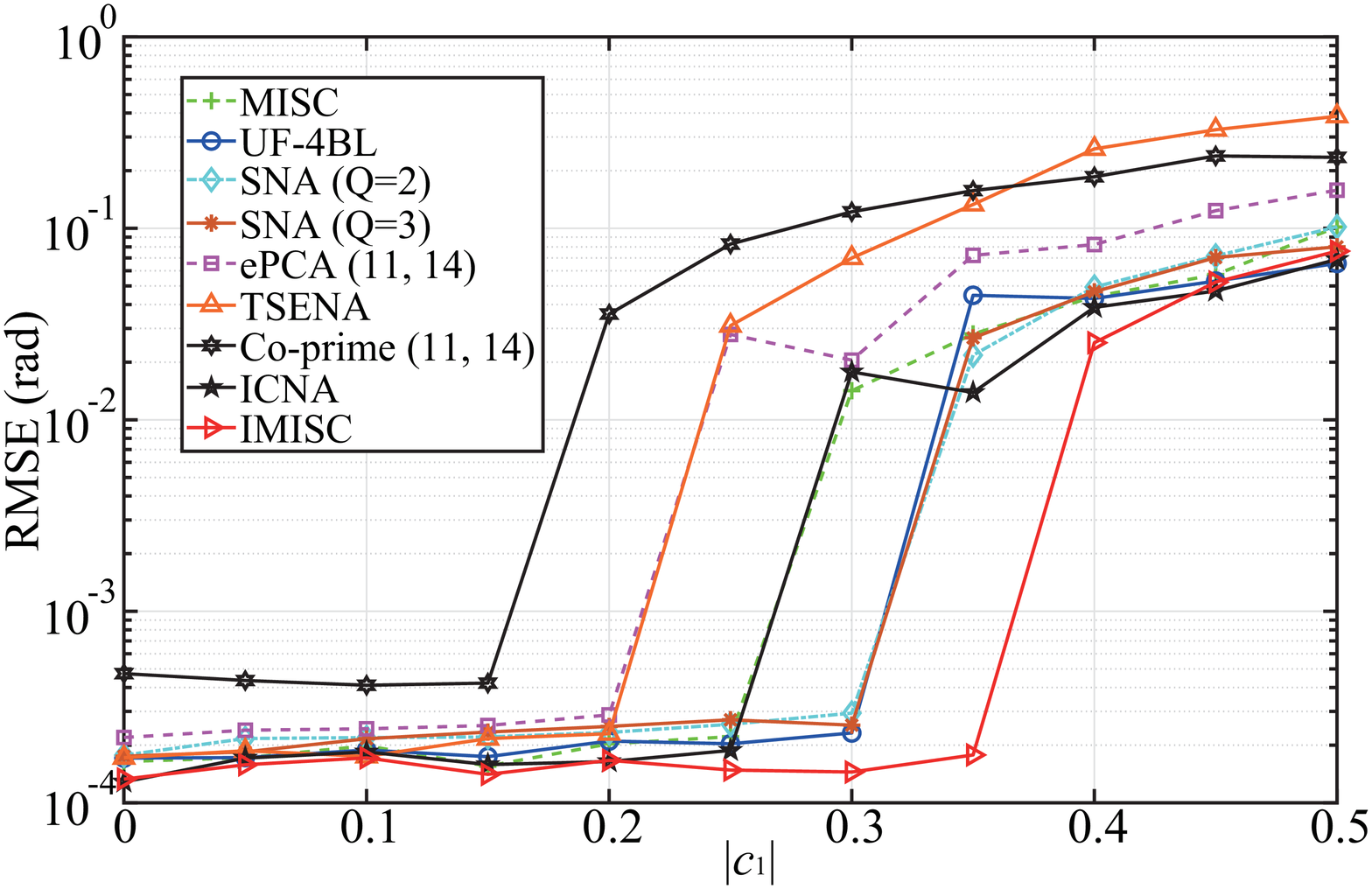} }\\
    \subfloat[RMSE versus snapshots.]{\includegraphics[width=0.5\columnwidth,height=3.3cm]{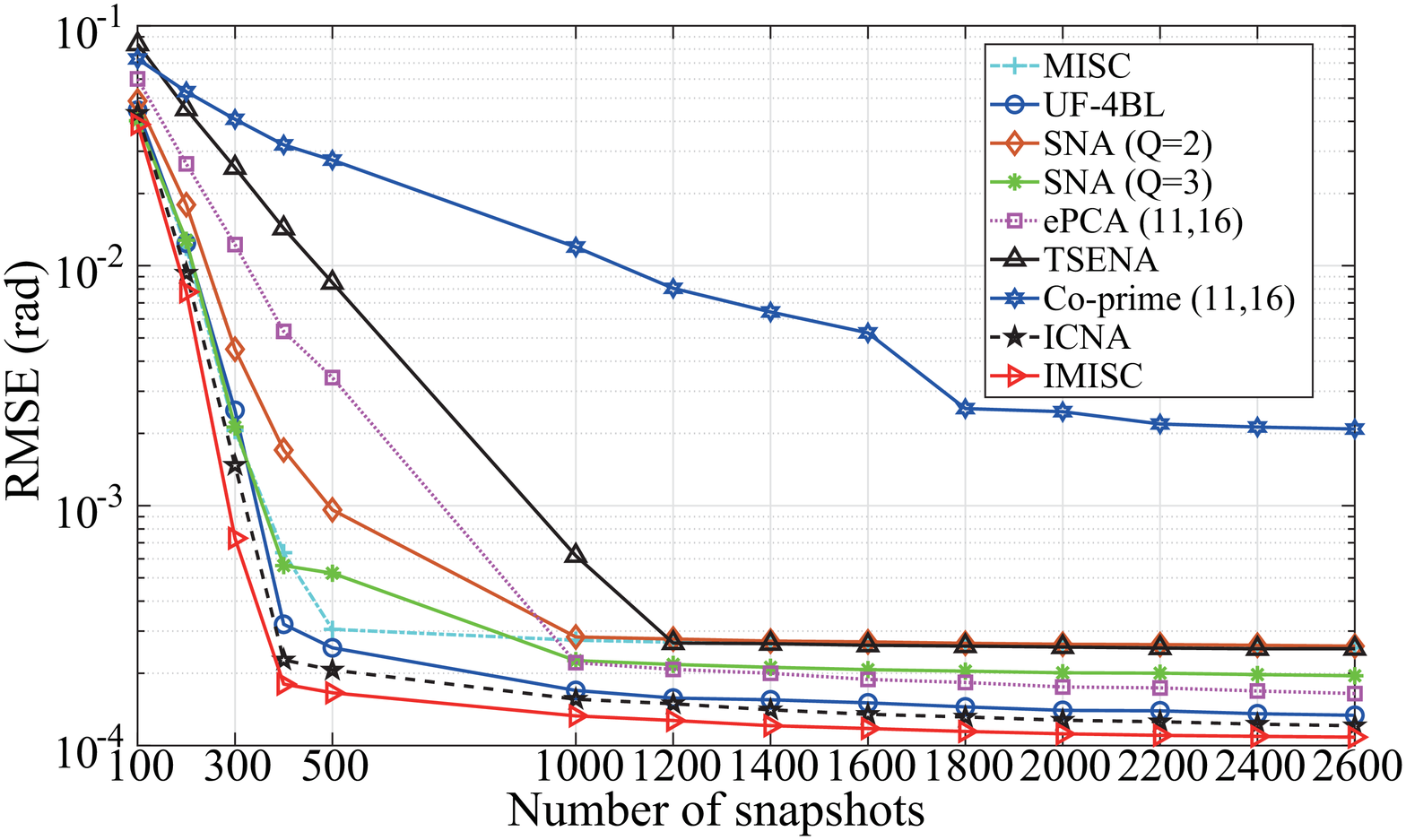} }
    \caption{RMSE performance in various conditions.}
    \label{SNR-C1}
\end{figure}

Firstly, the RMSE performance versus SNR in the presence of coupling is presented. All the SAs have 34 sensors and there are 39 sources in $[-60^\circ,60^\circ]$, and $a_1=0.3e^{j\pi/3}$. From~Fig.~\ref{SNR-C1}(a), it is observed that IMISC has the best performance, due to the longest uDOF and low MC.

Next, the RMSE versus $|a_1|$ is investigated. Here, all the SAs  consist of $35$ sensors, the number of sources is set to $50$ within azimuth $[-60^\circ,60^\circ]$, SNR=~0~dB, and $|a_1|$ varies from 0 to 0.5. As shown in~Fig.~\ref{SNR-C1}(b), when $|a_1|<0.4$, IMISC gives the best performance, while when $|a_1|>0.4$, the ICNA and UF-4BL perform better than IMISC.

{Finally, the RMSE performance versus the number of snapshots is presented in~Fig.~\ref{SNR-C1}(c), where all the SAs have~37 sensors, the number of sources is~45 within azimuth $[-60^\circ,60^\circ]$, SNR=0~dB, and $|a_1|=0.3$. We can see that the proposed IMISC provides the best performance over snapshots.}

\section{Conclusion}\label{conclusion}
In this letter, based on the MISC principle, a novel SA structure, termed as IMISC has been proposed. Compared with the MISC array the IMISC array possesses higher uDOF and lower coupling among sensors. The IMISC SA can be uniquely obtained by an inter-element spacing set, which is constrained by the maximum inter-element spacing and the number of sensors. Moreover, the uDOF and the weight function of the IMISC SAs are analyzed in detail. Simulations verify that IMISC can provide a great balance between uDOF and coupling leakage, and hence has better performance in DOA estimation than existing SAs. Future work will focus on how to further reduce the MC among sensors.

\begin{appendices}
\section{Consecutive part in the DCA of IMISC}\label{proof_proposition_SDCA_symmetric}
Based on~~(\ref{consecutive_part1}) and due to the symmetric property of DCA, we only need to prove $\mathbb{D}_{\text{IMISC}}^{+} \supset \mathbb{C}_{\text{IMISC}}^{+}$, where $\mathbb{D}_{\text{IMISC}}^{+}$ is the positive set of $\mathbb{D}_{\text{IMISC}}$ and $\mathbb{C}_{\text{IMISC}}^{+}=[1,MQ - \frac{{3{M^2}}}{4} - \frac{M}{2} + 1]$.
\par
In IMISC, assume that ULA~$i$ has the following position set
\begin{equation}
\begin{aligned}
\mathbb{P}_{\text{ULA }i}=\{p_{i}(x),x=1,\dots,Q_i\},
\end{aligned}
\label{ULAi}
\end{equation}
where $Q_i$ is the number of sensors in ULA~$i$. Similarly, setting the position set of ULA~$j$ as
\begin{equation}
\begin{aligned}
\mathbb{P}_{\text{ULA }j}=\{p_{j}(y),y=1,\dots,Q_j\},
\end{aligned}
\label{ULAj}
\end{equation}
where $Q_j$ is the number of sensors in ULA~$j$. Then, defining $\mathbb{D}_{i,j}$ as the DCA between ULA~$i$ and ULA~$j$, which has the following expression
\begin{equation}
\begin{aligned}
\mathbb{D}_{i,j}=\{p_{j}(y)-p_{i}(x),x=1,\dots,Q_i,y=1,\dots,Q_j\}.
\end{aligned}
\label{D_i_j}
\end{equation}
In particular, we have
\begin{equation}\nonumber
\begin{aligned}
\mathbb{D}_{i,i}=\{p_{i}(x),x=1,\dots,Q_i\}-p_{i}(1).
\end{aligned}
\label{D_i_i}
\end{equation}
In IMISC, the number of ULAs is~6, in this regard, $\mathbb{D}_{\text{IMISC}}^{+}$ can be expressed as
\begin{equation}
\begin{aligned}
\mathbb{D}_{\text{IMISC}}^{+}=\bigcup_{i=1}^{5} \mathbb{D}_{i,j},j=i,i+1,\dots,6,
\end{aligned}
\label{D_MISC+}
\end{equation}
where $\cup$ is the union operator. Based on~(\ref{Structure}) and~(\ref{D_i_j}), some DCA expressions in~(\ref{D_MISC+}) are provided in~(\ref{expressions_D}). The proof of $\mathbb{D}_{\text{IMISC}}^{+} \supset \mathbb{C}_{\text{IMISC}}^{+}$ can be completed by finding consecutive lags within the range $[1,MQ - \frac{{3{M^2}}}{4} - \frac{M}{2} + 1]$ based~(\ref{D_MISC+}) and~(\ref{expressions_D}) (at the bottom of this page).
\par
Based on~(\ref{expressions_D}), one can tell that $\mathbb{D}_{1,2}\cup\mathbb{D}_{1,3}\cup\mathbb{D}_{3,4}\cup\mathbb{D}_{4,5}\cup\mathbb{D}_{5,6}$ can generate consecutive range
\begin{equation}
[1,\frac{M^2}{8}-\frac{M}{4}],
\label{cons1}
\end{equation}
and $\mathbb{D}_{1,4}\cup\mathbb{D}_{2,4}\cup\mathbb{D}_{3,4}\cup\mathbb{D}_{4,5}\cup\mathbb{D}_{4,6}\cup\mathbb{D}_{4,4}$ can generate consecutive range
\begin{equation}
[\frac{M^2}{8}-\frac{M}{4}+1,MQ-\frac{7M^2}{8}-\frac{5M}{4}+3],
\label{cons2}
\end{equation}
and $\mathbb{D}_{1,5}\cup\mathbb{D}_{1,6}\cup\mathbb{D}_{2,5}\cup\mathbb{D}_{2,6}\cup\mathbb{D}_{3,5}\cup\mathbb{D}_{3,6}\cup\mathbb{D}_{1,4}\cup\mathbb{D}_{2,4}\cup\mathbb{D}_{4,6}$ can generate consecutive range
\begin{equation}
[MQ-\frac{7M^2}{8}-\frac{5M}{4}+3,MQ - \frac{{3{M^2}}}{4} - \frac{M}{2} + 1].
\label{cons3}
\end{equation}
Equations~(\ref{cons1}),~(\ref{cons2}) and~(\ref{cons3}) complete the proof.
\small
\hrule
\begin{align}\nonumber
&\mathbb{D}_{\text{1,2}}=\Big\{ 1,2,\dots,\frac{M}{2} \Big\},  \quad\mathbb{D}_{\text{2,6}}=\Big\{ MN-\frac{3M^2}{4}-M+1+i+j,i=0,1,j=0,2,\dots,\frac{M}{2}-2 \Big\}, \\\nonumber
&\mathbb{D}_{\text{1,3}}=\Big\{ \frac{M}{2}+i+j,i=0,2,\dots,\frac{M}{2}-2,j=(0,1,\dots,\frac{M}{4}-2)(\frac{M}{2}-1) \Big\},\\\nonumber
&\mathbb{D}_{\text{1,4}}=\Big\{ \frac{M^2}{8} + \frac{M}{4} + 2+i+j,i=0,2,\dots,\frac{M}{2}-2,j=(0,1,\dots,N-M-1)(M) \Big\},\\\nonumber
&\mathbb{D}_{\text{2,4}}=\Big\{ \frac{M^2}{8} + \frac{M}{4}+i+j,i=0,1,j=(0,1,\dots,N-M-1)(M) \Big\},\\\nonumber
&\mathbb{D}_{\text{3,4}}=\Big\{ M+i+j,i=(0,1,\dots,\frac{M}{4}-2)(\frac{M}{2}-1),j=(0,1,\dots,N-M-1)(M) \Big\},\\\nonumber
&\mathbb{D}_{\text{4,5}}=\Big\{ \frac{M}{2}+1+i+j,i=(0,1,\dots,\frac{M}{4}-2)(\frac{M}{2}+1),j=(0,1,\dots,N-M-1)(M) \Big\},\\\label{expressions_D}
&\mathbb{D}_{\text{4,6}}=\Big\{ \frac{M^2}{8} - \frac{M}{4} + 1+i+j,i=0,2,\dots,\frac{M}{2}-2,j=(0,1,\dots,N-M-1)(M) \Big\},\\\nonumber
&\mathbb{D}_{\text{5,6}}=\Big\{ 2+i+j,i=0,2,\dots,\frac{M}{2}-2,j=(0,1,\dots,\frac{M}{4}-2)(\frac{M}{2}+1) \Big\},\\\nonumber
&\mathbb{D}_{\text{1,5}}=\Big\{ MN-\frac{7M^2}{8}-\frac{M}{4}+3+i+j,i=0,2,\dots,\frac{M}{2}-2,j=(0,1,\dots,\frac{M}{4}-2)(\frac{M}{2}+1) \Big\},\\\nonumber
&\mathbb{D}_{\text{1,6}}=\Big\{ MN-\frac{3M^2}{4}-M+3+i+j,i=0,2,\dots,\frac{M}{2}-2,j=0,2,\dots,\frac{M}{2}-2 \Big\},\\\nonumber
&\mathbb{D}_{\text{2,5}}=\Big\{ MN-\frac{7M^2}{8}-\frac{M}{4}+1+i+j,i=0,1,j=(0,1,\dots,\frac{M}{4}-2)(\frac{M}{2}+1) \Big\},\\\nonumber
&\mathbb{D}_{\text{3,5}}=\Big \{MN-M^2+\frac{M}{2}+1+i+j,i=(0,1,\dots,\frac{M}{4}-2)(\frac{M}{2}-1),j=(0,1,\dots,\frac{M}{4}-2)(\frac{M}{2}+1)\Big\},\\\nonumber
&\mathbb{D}_{\text{3,6}}=\Big \{MN-\frac{7M^2}{8}-\frac{M}{4}+1+i+j,i=(0,1,\dots,\frac{M}{4}-2)(\frac{M}{2}-1),j=0,2,\dots,\frac{M}{2}-2 \Big\}.\\\nonumber
\end{align}\nonumber

\end{appendices}

\end{document}